%% file: main.tex
\documentclass{article}
\usepackage{kr} 

\usepackage{times}
\usepackage{soul}
\usepackage{url}
\usepackage[hidelinks]{hyperref}
\usepackage[utf8]{inputenc}
\usepackage[small]{caption}
\usepackage{graphicx}
\usepackage{amsmath}
\usepackage{amsthm}
\usepackage{amssymb}
\usepackage{booktabs}
\usepackage{algorithm}
\usepackage{algorithmic}
\usepackage{multirow}
\usepackage{tikz}
\usetikzlibrary{positioning}
\tikzset{>=stealth}
\usepackage{pgfplots}
\usepackage{subfigure}
\usepackage{enumitem}
\title{Revisiting Inferential Benchmarks for Knowledge Graph Completion}

\author{%
Shuwen Liu$^1$\and
Bernardo Cuenca Grau$^1$\and
Ian Horrocks$^1$\and
Egor V. Kostylev$^2$ \\
\affiliations
$^1$Department of Computer Science, University of Oxford, UK\\
$^2$Department of Informatics, University of Oslo, Norway\\
\emails
\{shuwen.liu, bernardo.cuenca.grau, ian.horrocks\}@cs.ox.ac.uk,
egork@ifi.uio.no
}

\begin{document}

\maketitle

\input{abstract}
\input{introduction}

\input{background}

\input{procedure}

\input{negative_sampling}
\input{evaluation}
\input{conclusion}

\section*{Acknowledgements}
 
This work was supported by the SIRIUS Centre for Scalable Data Access (Research
Council of Norway, project number 237889), and the EPSRC projects ConCur
(EP/V050869/1), UK FIRES (EP/S019111/1), and OASIS (EP/S032347/1). For the
purpose of Open Access, the author has applied a CC BY public copyright licence
to any AuthorAccepted Manuscript (AAM) version arising from this submission.





\appendix

\bibliographystyle{kr}
\bibliography{kr-sample}

\end{document}

%% file: abstract.tex
\begin{abstract}
Knowledge Graph (KG) completion is the problem of extending an incomplete KG with missing facts. 
A key feature of Machine Learning approaches for KG completion is 
their ability to learn inference patterns, so that the predicted facts are the results of applying these patterns to the KG. 
Standard completion benchmarks, however, are not well-suited for evaluating models' abilities to
learn patterns, because the training and test sets of these benchmarks are
a random split of a given KG and hence do not capture the causality of inference patterns.
We propose a novel approach for designing KG completion benchmarks based on the following principles:
 there is a set of logical rules 
so that the missing facts are the results of the rules' application;
the training set includes both premises matching rule antecedents and the corresponding conclusions; the test set consists of the results of applying the rules to the training set; 
the negative examples are designed to discourage the models from learning rules not entailed by the rule set.
We use our methodology to generate several benchmarks and evaluate a wide range of existing KG completion systems. Our results provide novel insights on the ability of existing models to induce inference patterns from
incomplete KGs.

\end{abstract}

%% file: introduction.tex
\section{Introduction}  
Knowledge Graphs (KGs) are graph-structured databases where nodes are entities
of interest and edges represent  
relations between such entities~\cite{DBLP:journals/csur/HoganBCdMGKGNNN21}. KGs
are commonly represented as a set of RDF triples~\cite{misc/manola04}, and
prominent KGs  in RDF format, such as DBpedia~\cite{DBLP:conf/semweb/AuerBKLCI07} and 
Freebase~\cite{DBLP:conf/sigmod/BollackerEPST08}, have been successfully exploited in
Web search, question answering, and recommendation tasks~\cite{DBLP:conf/emnlp/LuoLLZ18,DBLP:conf/www/WangZXLG19}.
These KGs are, however, highly incomplete; for example, over 93$\%$ of persons 
in Freebase have no place of birth, which is a compulsory attribute~\cite{DBLP:conf/naacl/MinGW0G13}.
This has motivated a growing interest in  \emph{KG completion}~\cite{DBLP:conf/nips/BordesUGWY13,DBLP:conf/esws/SchlichtkrullKB18,DBLP:conf/iclr/SunDNT19}: 
a learning task where
the aim is to extend  
a KG with missing triples that are likely to hold.
 
The availability of suitable benchmarks is key to the development of Machine Learning (ML) technologies, and a 
number of benchmarks such as FB15K-237~\cite{DBLP:conf/acl-cvsc/ToutanovaC15} and WN18RR~\cite{DBLP:conf/aaai/DettmersMS018} have become
de-facto standards for the evaluation of KG completion models. 
Positive examples in these benchmarks are obtained
by random splitting of the triples in a given real-life KG into training,  validation, and test sets; in turn,
negative examples are typically 
generated according to a specific
 \emph{negative sampling} method. 
 The most common method  is \emph{corruption}, where
 entities occurring in the subject (i.e., the first) or the object (i.e., the last) position in positive example triples
 are replaced with other entities sampled from the KG. 
The standard benchmarking approach based on random splitting 
is  well-suited for assessing the models' capability to learn a
(randomly generated) probability distribution on triples; however,
it also comes with significant shortcomings, which we discuss next.

 The first important limitation of standard benchmarks is that they provide little information on
 the models' ability to capture \emph{inference patterns}---that is, types of causal dependencies between premises and conclusions that may hold in the KGs~\cite{DBLP:conf/nips/AbboudCLS20}.
 Examples of such patterns include
\emph{symmetry} (e.g., if relation `is colleague' is symmetric and
$a$ is a colleague of $b$, then $b$ is also a colleague of $a$);
\emph{composition}  
(e.g., if `is grandmother of' is the composition of 
the relation `is mother of' with itself, $a$ is the mother of $b$, and $b$ is the mother of $c$, then $a$ is a grandmother of $c$); 
and \emph{intersection}
(e.g., if  `is mother of' is the intersection of relations `is parent of' and 'gives birth to', $a$ is a parent of $b$, and $a$ gives birth to $b$, then $a$ is the mother of $b$).
There is broad consensus that the ability of KG completion models to
learn inference patterns is key to improving the
reliability and explainability of their predictions~\cite{DBLP:series/ssw/BianchiRCPM20}.

Inference patterns can be formally represented as rule templates. 
For example,  the intersection pattern 
can be expressed using the following rule template, 
where  $\_\mathsf{R}$, $\_\mathsf{S}$ and $\_\mathsf{T}$ can be instantiated
to arbitrary relations:
\begin{equation}
\label{eq:ex_pattern}
(x, \_\mathsf{R}, y) \land (x, \_\mathsf{S}, y) \rightarrow (x, \_\mathsf{T}, y).
\end{equation}
In turn, instantiations of an inference pattern for specific relations
can be represented as Datalog rules. For instance, our example \eqref{eq:ex_pattern} of the intersection
pattern involving parenthood relations can be written as the following rule:  
\begin{equation}
\label{eq:ex_rule}
(x, \mathsf{IsParent}, y) {\land} (x, \mathsf{GivesBirth}, y) {\rightarrow} (x, \mathsf{IsMother}, y).
\end{equation}

The ability of KG completion systems to capture inference patterns has been recently analysed from a 
theoretical perspective~\cite{DBLP:conf/iclr/SunDNT19,DBLP:conf/nips/AbboudCLS20}.
For instance, it has been shown that RotatE~\cite{DBLP:conf/iclr/SunDNT19} is able to capture symmetry---that 
is, given a rule defining a relation as symmetric, there exists a configuration of the models' parameters associated to the relevant 
relation such that, for any dataset, the models' predictions 
will coincide with the facts derived by the symmetry rule;
in contrast, TransE~\cite{DBLP:conf/nips/BordesUGWY13} is not able to capture symmetry in this sense. 
These theoretical results, however, provide little indication of the models' capabilities to learn 
relevant patterns \emph{in practice}. To this end, there is a need for benchmarks
that take into account the causal dependencies inherent to inference patterns, thus satisfying the
following requirement.
\begin{enumerate}
    \item  During training, models should
    witness both the premise and conclusion triples 
    for selected rules instantiating a 
    pattern of interest;
    at the same time, triples used as positive validation and test examples 
    should have supporting
    evidence in the training set---that is, be the conclusions of 
    rule witnesses with premises in the
    training set.
    \end{enumerate}

The second important shortcoming of standard
benchmarks lies in their corruption-based strategy for
generating negative examples via sampling. As shown by Safavi and Koutra~\shortcite{DBLP:conf/emnlp/SafaviK20}, correct classification of corruption-based negative 
examples is nearly trivial for state-of-the-art KG completion systems. To further 
illustrate the limitations of corruption-based negative sampling, consider
the situation where the function to be learnt involves the dependency
formalised by rule  \eqref{eq:ex_rule}; however,  a
particular model learns instead the simpler rule
\begin{equation}
\label{eq:ex_rule1}
(x, \mathsf{IsParent}, y) \rightarrow (x, \mathsf{IsMother}, y).
\end{equation}
Rule \eqref{eq:ex_rule1} logically entails \eqref{eq:ex_rule}, but not vice-versa, and hence the model
can make (a potentially large number of) wrong predictions. 
Corruption-based negative sampling, however, would most likely not generate
negative examples that penalise the model for learning 
the unintended rule.
To address this issue, we need benchmarks where the negative examples (either included into the benchmark explicitly or produced by a sampling strategy) satisfy
the following requirement.
\begin{enumerate}
    \item[2.] Negative training examples should witness the rules that the model should not learn, in particular those that logically entail
    the rules selected for learning (but not the other way round); at the same time, the negative validation and test examples should also include such witnesses, so that a model is penalised for learning unintended rules.
 \end{enumerate}

Overall, we will call KG completion benchmarks satisfying Requirements
1 and 2 \emph{inferential benchmarks}, and in this paper we describe a principled 
approach for constructing such benchmarks of appropriate size and complexity.

\paragraph{Related Work.} 
Before describing our approach, we briefly 
discuss existing benchmarks based on inference patterns and argue that they
do not satisfy all our requirements for inferential benchmarks. Benchmarks Kinship~\cite{DBLP:conf/aaai/KempTGYU06} and Country~\cite{DBLP:conf/aaaiss/Bouchard0T15} are based on 
simple inference patterns and involve datasets of small size with a very limited number of relations.
Although these benchmarks satisfy our Requirement~1, they fail to satisfy Requirement 2 as
they do not include negative examples and (silently) rely on the corruption-based negative sampling strategy.
Cao et al.~\shortcite{DBLP:conf/acl/0002JLL0Z20} recently
proposed a more sophisticated benchmark InferWiki based on Wikidata. Their approach relies on the rule
mining system AnyBURL~\cite{DBLP:conf/ijcai/MeilickeCRS19} 
for generating relevant rules of a very specific syntactic shape.
Triples witnessing the premises of these rules 
are included in the training set as positive examples, 
while all the conclusion triples are included as positive examples in the test set; as a result,
InferWiki does not satisfy Requirement 1 since models are not able to witness during training both the 
premise and the conclusion triples for the selected rules.
Candidate negative examples are generated using
the conventional random corruption strategy and are then subsequently filtered by
human annotators on the base of their plausibility in real life; thus, InferWiki does not satisfy Requirement 2 either.

Several papers have already reported issues with existing KG completion benchmarks and evaluation protocols, including
the unchallenging nature of negative examples \cite{DBLP:journals/tkde/WangMWG17,DBLP:conf/cikm/BansalTR20,DBLP:conf/acl/0002JLL0Z20},
leakage of the test set~\cite{DBLP:conf/sigmod/AkramiSZHL20}, and the
potential unfairness of ranking-based metrics~\cite{DBLP:journals/corr/abs-2002-06914,DBLP:conf/acl/SunVSTY20}.

\paragraph{Our Contribution.}

In this paper, we propose a novel approach that enables researchers to create inferential benchmarks and assess the performance of their own KG completion systems. 
The pipeline of our approach for constructing inferential benchmarks 
starts with a KG and a set of inference patterns of interest, and consists of three main steps. The first step
generates rules for the selected inference patterns with large number of witnessing premises in the KG. 
The second step applies the rules and distributes the inferred triples amongst sets of positive examples
for training, validation, and testing, respectively, so that Requirement 1 is satisfied (the triples in the original KG are also taken as positive training examples).
Finally, the third step generates negative training, validation, and testing examples satisfying
Requirement 2 by means of one of three novel mathods.

Using our pipeline, we generated a collection of 
inferential benchmarks based on common inference patterns and the
KGs underpinning FB15K-237~\cite{DBLP:conf/acl-cvsc/ToutanovaC15}, 
WN18RR~\cite{DBLP:conf/aaai/DettmersMS018}, and LUBM~\cite{DBLP:journals/ws/GuoPH05}.
We then conducted a
comprehensive
evaluation
of KG completion systems on these benchmarks, including
 embedding-based TransE~\cite{DBLP:conf/nips/BordesUGWY13}, RotatE~\cite{DBLP:conf/iclr/SunDNT19}, ComplEx~\cite{DBLP:conf/icml/TrouillonWRGB16},
    DistMult~\cite{DBLP:journals/corr/YangYHGD14a}, and BoxE~\cite{DBLP:conf/nips/AbboudCLS20};
     GNN-based R-GCN~\cite{DBLP:conf/esws/SchlichtkrullKB18};
    and rule mining AnyBURL~\cite{DBLP:conf/ijcai/MeilickeCRS19} and
    RuleN~\cite{DBLP:conf/semweb/MeilickeFWRGS18}.
    
Our findings can be summarised as follows. 

\begin{itemize}
    \item All systems performed significantly worse on our benchmarks than on
    the standard ones, which suggests that benchmarks generated using our approach are
    challenging for state-of-the-art KG completion systems.  
    \item BoxE and RotatE are the best performing embedding-based models. Rule-based systems outperformed others
    on simple inference patterns, but not on  complex patterns.
    \item Some models achieved favourable performance despite their theoretical inability to  capture certain patterns.
       \item Performance on classification metrics relies heavily on the choice (or sampling strategy) of negative examples, with our generation methods leading to a considerable performance drop.
\end{itemize}
These findings highlight the benefits of our benchmarking approach and provide interesting
 insights for the further development of KG completion methods.

%% file: background.tex
\section{Background}\label{sec:background}
In this section, we first describe the basic notions underpinning
KGs and define the KG completion problem, then introduce the standard approach to KG completion benchmarking and the evaluation metrics they use, and finally 
recapitulate concepts related to Datalog and inference patterns.

\subsection{KGs and KG Completion}

In our context, a \emph{signature} consists of pairwise disjoint sets of \emph{types} and \emph{relations}, collectively referred to as \emph{predicates},
and \emph{constants}, which are also often referred to as \emph{entities}. A \emph{knowledge graph} (\emph{KG})
is a finite set of triples of the form $(e, \mathsf{type}, t)$, where $e$ is a constant and $t$ a type, and of the form $(s, R, o)$,  where $s$ and $o$ are constants,
and $R$ is a relation.
For $\mathcal{K}$ a KG, let
$\mathsf{Sig}(\mathcal{K})$, 
$\mathsf{Types}(\mathcal{K})$,
$\mathsf{Rels}(\mathcal{K})$,
$\mathsf{Preds}(\mathcal{K})$,
and 
$\mathsf{Consts}(\mathcal{K})$ denote
the signature,
the (set of) 
types, 
relations,
predicates,
and constants
used in $\mathcal{K}$, respectively. We write ${\mathsf{Sig}(\mathcal{K}) \subseteq \mathsf{Sig}(\mathcal{K}')}$ if the signature of $\mathcal K$ uses only predicates and constants in the signature of  $\mathcal{K}'$.

Intuitively,  \emph{KG completion} is the problem of extending
a KG to its complete version over the same signature.
It is customary to formalise this problem as a (transductive) ML
 task where, given an incomplete KG $\mathcal{K}$,
 the goal is to learn the Boolean \emph{completion function} 
 $f_{\mathcal K}(\cdot)$ applicable to triples over $\mathsf{Sig}(\mathcal{K})$ such that 
 $f_{\mathcal{K}} (\lambda)$ is true if  $\lambda$ is in the \emph{completion} $\mathcal{K}^*$ of $\mathcal K$. 
 The  \emph{confidence-based} variant of this task assumes that $f_{\mathcal K}(\lambda)$ is, for each
 triple $\lambda$, a value in $[0, 1]$ representing the likelihood that $\lambda$ holds in $\mathcal{K}^*$.

\subsection{KG Completion Benchmarks}\label{sec:def-benchmark}

Benchmarks play an important role in evaluating KG completion methods and thus in motivating further development of the field. 
KG completion benchmarks usually contain disjoint sets $\mathcal{P}_\text{train}$, $\mathcal{P}_\text{valid}$ (possibly), and $\mathcal{P}_\text{test}$ of triples
for training, validation, and testing, respectively, with ${\mathsf{Sig}(\mathcal{P}_\text{valid}) \subseteq \mathsf{Sig}(\mathcal{P}_\text{train})}$ and ${\mathsf{Sig}(\mathcal{P}_\text{test}) \subseteq \mathsf{Sig}(\mathcal{P}_\text{train})}$. Let $\mathcal{P}_\text{all}$ denote the union of these three sets. 
Triples $\mathcal{P}_\text{train}$, $\mathcal{P}_\text{valid}$, and $\mathcal{P}_\text{test}$ are assumed to be positive training, validation, and test examples; so, each KG completion benchmark 
may also contain sets $\mathcal{N}_\text{train}$, $\mathcal{N}_\text{valid}$, and $\mathcal{N}_\text{test}$ of negative 
examples for training, validation, and testing, respectively~\cite{DBLP:conf/emnlp/SafaviK20,DBLP:conf/acl/0002JLL0Z20}. 
It is common, however,  not to include explicit negative examples into a benchmark, but instead rely on a sampling strategy for generating
these examples~\cite{DBLP:conf/nips/SocherCMN13}, thus
adopting a (partial) \emph{closed-world} assumption---that is, assuming the triples not observed in $\mathcal{P}_\text{all}$ to be false. It is 
worth to note here that taking all unobserved triples over the signature as negative examples instead of sampling not only 
leads to unacceptable imbalance between positive and negative examples, 
but is also computationally prohibitive.

The most common negative sampling strategy is to randomly \emph{corrupt} one of the three components 
of a positive example triple~\cite{DBLP:conf/nips/BordesUGWY13}.
For instance, for a positive test example $(s, R, o)$, the triples $(s', R, o)$, $(s, R, o')$, and $(s, R', o)$ may be taken as negative test examples, where $s'$ and 
$o'$ are randomly sampled from $\mathsf{Consts}(\mathcal{P}_\text{train})$ and $R'$ is randomly sampled from $\mathsf{Rels}(\mathcal{\mathcal{P}_\text{train}})$ 
so that the resulting triple is not in $\mathcal{P}_\text{all}$ (it is also not required that exactly 
three corrupted triples are constructed for each positive example, and variations are possible; for example, corrupting $R$ is less common, while  $s$ and $o$ are often corrupted a given number of times). 
Note, however, that negative examples generated in this way can be easily predicted as false, which often leads to nearly perfect performance~\cite{DBLP:conf/emnlp/SafaviK20}. 
In Section~\ref{sec:negative-sampling}, we will discuss alternative methods for generating more challenging negative examples.

KG completion systems are evaluated on benchmarks
using \emph{classification-based} and \emph{ranking-based} metrics, which are both
computed based on their predictions on positive $\mathcal{P}_\text{test}$ and negative $\mathcal{N}_\text{test}$ test examples (where $\mathcal{N}_\text{test}$ is either given by a benchmark or sampled as explained above).
The basic classification metrics are based on the counts $tp$, $tn$, $fp$, and $fn$ of true positive, true negative, false positive, and false negative predictions, respectively, 
which are computed in the usual way from $\mathcal{P}_\text{test}$ and $\mathcal{N}_\text{test}$, and the predictions of a model trained on $\mathcal{P}_\text{train}$ and $\mathcal{N}_\text{train}$, and validated on $\mathcal{P}_\text{valid}$ and $\mathcal{N}_\text{valid}$ 
(systems designed for the confidence-based variant of the completion task usually rely on a threshold hyperparameter to obtain a Boolean prediction).  Standard
classification-based metrics include \emph{precision} $Prec = tp/(tp+fp)$, \emph{recall} $Rec = tp/(tp+fn)$, \emph{accuracy} $Acc = (tp + tn) / (tp + tn + fp + fn)$, and
 \emph{F1} $F1 = 2 * Prec * Rec /(Prec + Rec)$. For confidence-based systems, the \emph{Receiver Operator Characteristic Area Under the Curve} (\emph{ROC AUC}) is also a commonly used metric~\cite{DBLP:journals/pr/Bradley97}.

Ranking-based metrics rely on confidence predictions in $[0, 1]$ and hence are
applicable only to systems solving the confidence-based variant of the completion task. 
These metrics usually take into account only triples of the form $(s, R, o)$, but 
a generalisation to triples $(c, \mathsf{type}, t)$ is straightforward.
For each $\lambda = (s, R, o)$ in $\mathcal{P}_\text{test}$, let $\mathcal{N}_\lambda^\mathsf{s}$ be the subset of $\mathcal{N}_\text{test}$ of all triples of the form $(s', R, o)$ (i.e., $\lambda$ with corrupted $s$), and let $\mathcal{N}_\lambda^\mathsf{R}$ and $\mathcal{N}_\lambda^\mathsf{o}$ be computed analogously.
Then, for each $x \in \{\mathsf{s}, \mathsf{R}, \mathsf{o}\}$, let $rank_x(\lambda)$ be the position of $\lambda$ in the ordering of ${\{\lambda\} \cup \mathcal{N}_\lambda^x}$ based on the prediction confidences
of the model (trained on $\mathcal{P}_\text{train}$, $\mathcal{N}_\text{train}$ and validated on $\mathcal{P}_\text{valid}$, $\mathcal{N}_\text{valid}$).
The \emph{constant-hit} $\text{C-Hits}@k$ and \emph{relation-hit} $\text{R-Hits}@k$ metrics for a number $k \in \mathbb{N}$ are then defined as ${\text{C-Hits}@k = (\text{Hits}_\mathsf{s}@k + \text{Hits}_\mathsf{o}@k)/2}$ and $\text{R-Hits}@k = \text{Hits}_\mathsf{R}@k$, respectively, where, for every $x$,  
${\text{Hits}_x@k = |\{ \lambda \in \mathcal{P}_\text{test}\,|\,rank_x(\lambda) \leq k \}| / | \mathcal{P}_\text{test}|}$. 
Furthermore, the \emph{Mean Reciprocal Rank} (\emph{MRR})
\emph{for constants} and \emph{relations} are defined as $\text{C-MRR} = (\text{MRR}_\mathsf{s} + \text{MRR}_\mathsf{o})/2$ and $\text{R-MRR} = \text{MRR}_\mathsf{R}$, respectively, where, for each $x$,  
$\text{MRR}_x = \left(\sum_{\lambda \in \mathcal{P}_\text{test}}\frac{1}{rank_x(\lambda)}\right) \Big/ ~\vert \mathcal{P}_\text{test}\vert$.

\subsection{Datalog}\label{sec:def-inference}

Our benchmark construction for KG completion relies on the concept of \emph{inference pattern},
which is an abstraction of a set of logical rules describing a type of causal dependencies that may exist in the KG.
The concrete logical rules included in the benchmarks are represented in Datalog, a well-known rule language for knowledge representation.

In our context, a (Datalog) \emph{atom} is an expression of the form  $(d, \mathsf{type}, t)$, $(d_1, R, d_2)$,  or $(d_1 \neq d_2)$ where $t$ is a type, $R$ is a relation, and each of $d$, $d_1$ and $d_2$ is either a constant or a \emph{variable}.
A (Datalog) \emph{rule} is a function-free first-order logic sentence of the form
\begin{equation}
B_1 \land \dots \land B_n \rightarrow H, \label{eq:def-rule}
\end{equation}
where $H$ is a $\neq$-free atom, which is called the \emph{head} of the rule, all $B_i$ are atoms serving together as the \emph{body} of the rule, and each variable in the rule is mentioned in some $\neq$-free $B_i$.
A (Datalog) \emph{program} is a finite set of rules.

A \emph{substitution} is an assignment of
constants to variables, and it extends to atoms and conjunctions of atoms in the usual way.
Each rule $r$ of form \eqref{eq:def-rule} realises a \emph{one-step rule application} $\mathcal{T}_r$ on KGs: for a KG $\mathcal{K}$,
KG $\mathcal{T}_r(\mathcal{K})$ consists of all triples $\sigma(H)$ for all \emph{witnesses} of the body in $\mathcal K$---that is, substitutions $\sigma$ such that $\sigma(d_1) \neq \sigma(d_2)$ for each $B_i$ of the form $(d_1 \neq d_2)$ and $\sigma(B_i) \in \mathcal K$ for each other $B_i$.
We call all such $\neq$-free $\sigma(B_i)$ in $\mathcal K$ the \emph{premise triples}
for $r$ in $\mathcal{K}$ and all such $\sigma(H)$ the \emph{conclusion triples} for $r$ in $\mathcal{K}$ (note that a triple may be both a premise and conclusion triple at the same time); moreover, we call all such $\sigma$ the \emph{support} of $r$ in $\mathcal K$.
The \emph{one-step application} $\mathcal{T}_{\mathcal R}(\mathcal{K})$ of a program $\mathcal R$ to a KG $\mathcal K$ is defined as $\mathcal{T}_{\mathcal R}(\mathcal{K}) = \bigcup_{r \in \mathcal R}\mathcal{T}_r(\mathcal{K})$. 
\emph{Materialisation} (or \emph{forward chaining}) is a reasoning paradigm which consists of successive rounds of one-step rule applications until no new triples can be derived for an input program and a KG~\cite{DBLP:journals/ai/MotikNPH19}.
For a program $\mathcal R$ and a KG $\mathcal{K}$,  the \emph{materialisation} $\mathcal{M}_{\mathcal R}(\mathcal{K})$ of $\mathcal R$ on $\mathcal{K}$ is defined as $\mathcal{M}_{\mathcal R}(\mathcal{K}) = \bigcup_{i \geq 0} \mathcal{K}_i$, where $\mathcal{K}_0 = \mathcal{K}$ and $\mathcal{K}_{i+1} = \mathcal{T}_{\mathcal R}(\mathcal{K}_i) \cup \mathcal{K}_i$ for each $i \geq 0$. 
A set $\mathcal R$ of rules \emph{logically entails} another set $\mathcal R'$, written $\mathcal R \models \mathcal R'$, if $\mathcal{M}_{\mathcal R'}(\mathcal{K}) \subseteq\mathcal{M}_{\mathcal R}(\mathcal{K})$ for each KG $\mathcal K$.

An \emph{inference (rule) pattern} is an expression of the form
\begin{equation}\label{eq:pattern}
\mathfrak{B}_1 \land \cdots \land \mathfrak{B}_n \rightarrow \mathfrak{H},
\end{equation}
where the $\mathfrak{B}_i$ and $\mathfrak{H}$  are same as the $B_i$ and $H$ in rule~\eqref{eq:def-rule}, respectively, except that instead of types $t$ and relations $R$ they use \emph{type templates} $\_\mathsf{t}$ and \emph{relation templates} $\_\mathsf{R}$, which are coming from dedicated infinite sets of symbols; it is worth emphasising that the template in $\mathfrak{H}$ may or may not be mentioned in one or several $\mathfrak{B}_i$. 
A rule $r$ is \emph{represented} by a inference pattern $\mathfrak{p}$ if $r$ can be obtained from $\mathfrak{p}$ by
substituting type and relation templates by types and relations, respectively.
In principle, our benchmarking approach works for arbitrary inference patterns; 
however, most of our concrete benchmarks (see Section~\ref{sec:evaluation}) rely on the common patterns
summarised in Table~\ref{tab:patterns}.

%% file: procedure.tex
\section{Inferential Benchmark Construction}\label{sec:procedure}
In this section, we introduce our pipeline for constructing KG completion benchmarks 
satisfying the two key requirements postulated in the  introduction.
An inferential benchmark is built from a KG $\mathcal{K}$ 
and one or several inference patterns $\mathfrak{P}$.
Our approach for constructing an inferential benchmark for $\mathcal{K}$ and  $\mathfrak{P}$ then consists of three steps:

\input{figure_procedure}

\begin{enumerate}
\item rule generation, which produces a set $\mathcal R$ of rules based on $\mathcal{K}$ and $\mathfrak{P}$ so that the size of their support (i.e., the number of witnessing premises) in $\mathcal{K}$ is maximised;

\item rule application and distributing the results, which distributes the one-step application result $\mathcal{T}_{\mathcal R}(\mathcal{K})$ and $\mathcal K$ itself into the training, validation, and test positive sets $\mathcal{P}_\text{train}$, $\mathcal{P}_\text{valid}$, and $\mathcal{P}_\text{test}$, so that Requirement 1 is satisfied;
\item negative example generation, which constructs appropriate and challenging negative examples according to one of three methods, so that Requirement 2 is satisfied.
\end{enumerate}
A summary of our pipeline 
is depicted in Fig.~\ref{fig:procedure}(a), and a simple example benchmark is given
in Fig.~\ref{fig:procedure}(b).

\subsection{Rule Generation}\label{sec:rule-generation}

\input{table_patterns}

We initiate rule generation by constructing a set $\mathcal{R}_\text{cand}$ of candidate rules from inference patterns $\mathfrak{P}$ (and  $\mathsf{Preds}({\mathcal{K}})$). 
First, for each pattern in $\mathfrak{P}$ with the head predicate (type or relation) template mentioned in the body, $\mathcal{R}_\text{cand}$ contains each
 rule  obtained from the pattern by substituting all predicate templates by predicates in $\mathsf{Preds}({\mathcal{K}})$. 
 Second,  
 for each pattern in $\mathfrak{P}$ with the head  template not mentioned in the body, $\mathcal{R}_\text{cand}$ contains 
 one rule for every substitution of the templates in the body as above; in turn, the head template is substituted by a random type or relation from $\mathsf{Types}({\mathcal{K}})$ and $\mathsf{Rels}({\mathcal{K}})$, respectively. 
 This is justified by the fact that rules differing only in the head predicate are essentially equivalent for learning purposes.

Although set $\mathcal{R}_\text{cand}$ may be  very  large, 
the majority of its rules are not useful for generating examples as they
do not apply to $\mathcal K$ sufficiently many times. So, we complete the rule generation step by selecting
 a subset $\mathcal R$ of rules $\mathcal{R}_\text{cand}$ with large support
 in $\mathcal K$. One approach is to select into $\mathcal R$ a fixed predefined 
 number of relations with the largest support; note, however, that we do this 
 separately for each pattern to ensure that each of them is represented 
 and can be learned---that is, for eack pattern, we include $k_1$ rules with the largest support in $\mathcal K$, where $k_1$ is a pre-defined number that can be 
 customised based on the expected size and rule diversity of the benchmark. 
The support can be computed using a SPARQL engine (we used RDFox~\cite{DBLP:conf/semweb/NenovPMHWB15}, see Section~\ref{sec:evaluation}). 
Selecting rules with the largest support is, however, not essential, and
an alternative approach is to manually select rules with large enough support. We will use both approaches in our benchmarks.

Instead of using rule mining models to generate rules from the KG, which only generate very specific types of rules, we choose to generate rules more randomly. As a result, we may 
generate rules
 that do not make sense from a modelling perspective,
such as ${(x, \mathsf{Speaks}, y) \rightarrow (x, \mathsf{LocatedIn}, y)}$.
This is justified by that fact that our aim is to design benchmarks that test the ability to learn rules according to 
inference patterns, rather than 
according to modelling considerations.

\subsection{Distributing Rule Application Results}\label{sec:rule-application}

The second step constructs the positive examples in
$\mathcal{P}_\text{train}$, $\mathcal{P}_\text{valid}$, and 
$\mathcal{P}_\text{test}$ for training, validation, and testing, respectively. 
To satisfy our Requirement 1,  
we include the original KG $\mathcal K$ in $\mathcal{P}_\text{train}$ and then
distribute the triples in $\mathcal{T}_{\mathcal R}(\mathcal{K})$, obtained by
applying the selected rules $\mathcal R$ to $\mathcal K$, between the 
three sets $\mathcal{P}_\text{train}$, $\mathcal{P}_\text{valid}$, and 
$\mathcal{P}_\text{test}$ as described next.

To avoid data leakage (i.e., the situation where
test examples are observed during training), we only consider the newly derived triples for distribution. Moreover, 
we distribute these triples independently rule by rule to ensure that 
each rule can be learned, at the same time ensuring that the same triple does 
not end up in more than one of the three sets (this must be done with
care since a triple may be derived by several different rules);
however, to ensure a reasonable size of the dataset, we sample a fixed number of triples for each of the rules before distribution.

We first compute ${\mathcal{T}_r(\mathcal{K}) \setminus \mathcal K}$ for each rule $r \in \mathcal R$, then randomly sample up to $k_2$ 
triples from ${\mathcal{T}_r(\mathcal{K}) \setminus \mathcal K}$ (where $k_2$ is again a pre-defined number that can be 
 customised based on the expected size), and split
the sampled triples into three sets,
$\mathcal{P}^r_\text{train}$, $\mathcal{P}^r_\text{valid}$, $\mathcal{P}^r_\text{test}$, according to a predefined ratio (e.g., 8:1:1 in our benchmarks, see Section~\ref{sec:evaluation}). Then, we take 
$$
\mathcal{P}_\text{train} =   \left(\bigcup\nolimits_{r\in \mathcal{R}}{\mathcal{P}_\text{train}^r}\right) \cup \mathcal{K}
$$ 
as the positive training set, 
$$
\mathcal{P}_\text{valid} =   \left(\bigcup\nolimits_{r\in \mathcal{R}}{\mathcal{P}_\text{valid}^r}\right)\setminus \mathcal{P}_\text{train}
$$ 
as the positive example triples for validation, and 
$$
\mathcal{P}_\text{test} =   \left(\bigcup\nolimits_{r\in \mathcal{R}}{\mathcal{P}_\text{test}^r}\right) \setminus \left(\mathcal{P}_\text{train} \cup \mathcal{P}_\text{valid}\right)
$$ 
as the positive example triples for testing. Finally, we again let $\mathcal{P}_\text{all} = \mathcal{P}_\text{train} \cup \mathcal{P}_\text{valid} \cup \mathcal{P}_\text{test}$.

%% file: figure_procedure.tex
\begin{figure*}[t]
\subfigure[\hspace{-6pt}]{
\begin{minipage}[t]{0.64\textwidth}
\centering
\begin{tikzpicture}[every node/.style={scale=0.6}]
\usetikzlibrary{shapes}
\tikzstyle{existing_user} = [draw, circle]
\tikzstyle{new_user} = [draw, circle, dashed]
\tikzstyle{existing_node} = [draw, rectangle]
\tikzstyle{item} = [align=left, fill=white]
\tikzstyle{rule} = [draw, rectangle]
\tikzstyle{rule2} = [draw, dotted, thick, rectangle]
\tikzstyle{rule3} = [draw, ellipse]
\tikzstyle{new_node} = [draw, rectangle, dashed]
\tikzstyle{known_edge} = [draw,->, thick]
\tikzstyle{predicted_edge} = [draw,->, dotted, thick]

\foreach \pos/\name/\label in {{(-1.9,0.35)/dataset1/KG $\mathcal{K}$}, {(1.02,0.7)/dataset2/KG $\mathcal{K}$},
{(6.1,1.2)/dataset3/$\mathcal{K}$}},
    \node[rule] (\name) at \pos {\label};
\foreach \pos/\name/\label in {{(-1.9,-1)/patterns/Inference patterns $\mathfrak{P}$ }},
    \node[rule3][align=center, text width=1.8cm] (\name) at \pos {\label};
\node[draw, dotted, thick, rectangle,align=center, text width=1.2cm] (new) at (3.6,-0.3) {$\mathcal{T}_\mathcal{R}(\mathcal{K})$};
\foreach \pos/\name/\label in {
{(5,1.2)/train/$\left(\bigcup\nolimits_{r\in \mathcal{R}}{\mathcal{P}_\text{train}^r}\right)$}, {(5.35,-0.3)/valid/$\left(\bigcup\nolimits_{r\in \mathcal{R}}{\mathcal{P}_\text{valid}^r}\right)\setminus \mathcal{P}_\text{train}$},
{(5.1,-1.8)/test/$\left(\bigcup\nolimits_{r\in \mathcal{R}}{\mathcal{P}_\text{test}^r}\right) \setminus \left(\mathcal{P}_\text{train} \cup \mathcal{P}_\text{valid}\right)$}},
    \node[rule2] (\name) at \pos {\label};
\foreach \pos/\name/\label in {{(1,-0.3)/rules/Rule set $\mathcal{R}$ }},
    \node[rule3][align=center, text width=1.2cm] (\name) at \pos {\label};

\foreach \pos/\name/\label in {{(7.88,1.2)/training_set/Positive training set $\mathcal{P}_\text{train}$},
{(7.88,0.3)/training_set2/Negative training set $\mathcal{N}_\text{train}$},
{(8.05,-0.3)/validation_set/Positive validation set
$\mathcal{P}_\text{valid}$},
{(8.05,-1.2)/validation_set2/Negative validation set $\mathcal{N}_\text{valid}$},
{(7.95,-1.8)/test_set/Positive test set
$\mathcal{P}_\text{test}$},
{(7.95,-2.7)/test_set2/Negative test set $\mathcal{N}_\text{test}$}},
    \node[rule] (\name) at \pos {\label};

\foreach \pos/\name/\label in {{(-1.9,-0.3)/plus/+}, 
{(1.02,0.3)/plus2/+},{(-1.2,-0.3)/start/$ $}, {(0.3,-0.3)/end/$ $},
{(1.7,-0.3)/start2/$ $}, {(3.1,-0.3)/end2/$ $},
{(-0.5,0.5)/rule/Rule}, {(-0.5,0.2)/generation/generation}, 
{(2.3,0.5)/rule2/Rule}, {(2.3,0.2)/application/application}, {(5.8,1.2)/plus3/+},
{(6.5,1.2)/eq1/$=~$},
{(6.58,-0.3)/eq2/$=~$},
{(6.75,-1.8)/eq3/$=~$}}
    \node[item] (\name) at \pos {\label};
\foreach \source/ \dest / \weight in 
{
start/end/$ $,
start2/end2/$ $,
new/train/$ $,
new/valid/$ $,
new/test/$ $
}
    \path[known_edge] (\source) -- node[] {$\footnotesize\text{\weight}$} (\dest);
\foreach \source/ \dest / \weight in 
{training_set/training_set2/Negative example generation,
validation_set/validation_set2/Negative example generation,
test_set/test_set2/Negative example generation
}
    \path[known_edge] (\source) -- node[fill=white!20] {$\footnotesize\text{\weight}$} (\dest);
\end{tikzpicture}
\end{minipage}
}
\subfigure[\hspace{-6pt}]{\
\begin{minipage}[t]{0.4\textwidth}
\centering
\begin{tikzpicture}[every node/.style={scale=0.6}]
\tikzstyle{vector} = [fill=white,align=center]
\tikzstyle{item2} = [align=left, fill=white]
\tikzstyle{arrow} = [draw,->,thick]
\node[vector] at (0,5) {
$\begin{aligned}
\text{Positive training set:}\\
(\text{Alex}, \mathsf{IsColleague}, \text{Bob}),
(\text{Bob}, \mathsf{IsColleague}, \text{John}),\\
(\text{Alex}, \mathsf{IsColleague}, \text{John}),\\
(\text{Harry}, \mathsf{IsColleague}, \text{James}),
(\text{James}, \mathsf{IsColleague}, \text{Tony}),\\
(\text{Harry}, \mathsf{IsColleague}, \text{Tony}),\\
(\text{Ada}, \mathsf{IsColleague}, \text{Eve}),
(\text{Eve}, \mathsf{IsColleague}, \text{Lucy})\;\\
\text{Negative training set:}\\
(\text{Alex}, \mathsf{IsColleague}, \text{James}),
(\text{Alex}, \mathsf{IsColleague}, \text{Tony}),\\
(\text{Alex}, \mathsf{IsColleague}, \text{Eve}),
(\text{Harry}, \mathsf{IsColleague}, \text{John}),\\
(\text{James}, \mathsf{IsColleague}, \text{John}),
(\text{Ada}, \mathsf{IsColleague}, \text{John}),\\
(\text{Harry}, \mathsf{IsColleague}, \text{Eve}),
(\text{Harry}, \mathsf{IsColleague}, \text{Lucy})\;\\
\text{Positive test set:}\\
(\text{Ada}, \mathsf{IsColleague}, \text{Lucy})\;\\
\text{Negative test set:}\\
(\text{Ada}, \mathsf{IsColleague}, \text{James})\;\\
\end{aligned}$};
\draw [thick,color=gray!50, dotted] (-2.4,6.92) rectangle (2.4, 5.45);
\draw [thick,color=gray!50, dotted] (-2.4,5.15) rectangle (2.4,3.95);
\draw [thick,color=gray!50, dotted] (-2.4,3.7) rectangle (2.4,3.35);
\draw [thick,color=gray!50, dotted] (-2.4,3.1) rectangle (2.4,2.75);
\end{tikzpicture}
\end{minipage}
}
\caption{\textnormal{(a)  Summary of our pipeline for constructing an example inferential benchmark~~ 
(b) An example inferential benchmark   
based on the rule $(x, \mathsf{IsColleague}, y) \land (y, \mathsf{IsColleague}, z) \rightarrow (x, \mathsf{IsColleague}, z)$ 
and the position-aware corruption method for negative example generation (validation set is omitted for simplicity); 
during training, a model could witness 
$(\text{Alex}, \mathsf{IsColleague}, \text{Bob}), (\text{Bob}, \mathsf{IsColleague}, \text{John})$ as premises, and 
$(\text{Alex}, \mathsf{IsColleague}, \text{John})$ as conclusion, and similar witness is enabled for Harry, James, and Tony;
for the positive test example $(\text{Ada}, \mathsf{IsColleague}, \text{Lucy})$, its premises $(\text{Ada}, \mathsf{IsColleague}, \text{Eve})$ and $(\text{Eve}, \mathsf{IsColleague}, \text{Lucy})$ are included in the training set
}}
\label{fig:procedure}
\end{figure*}
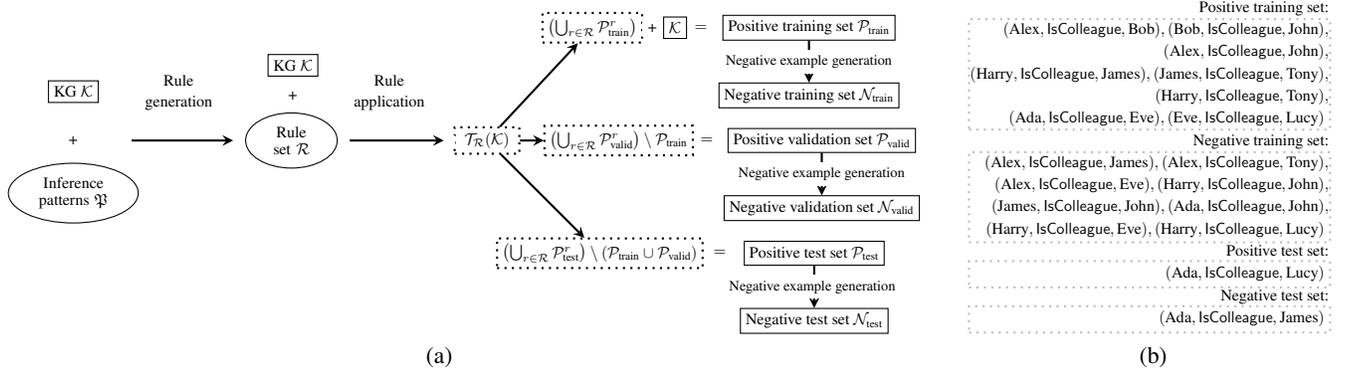

%% file: table_patterns.tex
\begin{table}
\begin{small}
\center
\renewcommand\arraystretch{1}
\setlength{\tabcolsep}{1mm}{
\begin{tabular}{ll}
\toprule
    & Pattern\\
\midrule
Symmetry   & $(x, \_\mathsf{R}, y) \rightarrow (y, \_\mathsf{R}, x)$ \\
Inversion   &$(x, \_\mathsf{R}, y) \rightarrow
(y, \_\mathsf{S}, x)$\\
Hierarchy & $(x, \_\mathsf{R}, y) 
        \rightarrow (x, \_\mathsf{S}, y)$\\
\midrule   
Composition   &$(x, \_\mathsf{R}, y) \land 
        (y, \_\mathsf{S}, z)
        \rightarrow (x, \_\mathsf{T}, z)$\\
Intersection  & $(x, \_\mathsf{R}, y) \land 
        (x, \_\mathsf{S}, y)
        \rightarrow (x, \_\mathsf{T}, y)$ \\
\midrule
\multirow{2}{*}{Triangle}  & $(x, \_\mathsf{R}, y)\land (x, \_\mathsf{S}, z) \land (y, \_\mathsf{T}, z) \land (x \neq y) $ \\ 
 &$~~~~{} \land (x \neq z) \land (y \neq z) \rightarrow (x, \_\mathsf{P}, y)$ \medskip\\
\multirow{3}{*}{Diamond} & $(x, \_\mathsf{R}, y)\land (x, \_\mathsf{S}, z) \land (y, \_\mathsf{T}, w) \land (z, \_\mathsf{P}, w)  $\\
&$~~~~{} \land (x \neq y) \land (x \neq z) \land (x \neq w)$
$ \land (y \neq z) $\\
& $~~~~{} \land (y \neq w) \land (z \neq w) \rightarrow (x, \_\mathsf{Q}, y)$\\
\bottomrule
\end{tabular}
}
\caption{Inference patterns considered in this paper}
\label{tab:patterns}
\end{small}
\end{table}

%% file: negative_sampling.tex
\subsection{Negative Example Generation}
\label{sec:negative-sampling}

The third step of our approach tackles the generation of
negative examples $\mathcal{N}_\text{train}$, $\mathcal{N}_\text{valid}$, and $\mathcal{N}_\text{test}$ for training, validation, and testing, respectively, 
so that Requirement 2 for inferential benchmarks is satisfied. In particular,
negative examples are generated so as to witness rules that the model should not learn, especially those
that logically entail rules from $\mathcal R$ (but are not equivalent to any rule in $\mathcal R$). In contrast to standard benchmarks, which  provide
a (corruption-based) negative sampling strategy, benchmarks produced by our approach do 
include concrete negative examples; this allows us to specify more precisely the undesired dependencies that should be prevented from learning, and hence 
make negative examples more challenging to classify.

We introduce three methods for generating negative examples: relevance-based sampling, position-aware corruption, and query-guided sampling. 
All three methods ensure balance between positive and negative examples---that is,  $|\mathcal{N}_\text{train}| = |\mathcal{P}_\text{train}|$, $|\mathcal{N}_\text{valid}| = |\mathcal{P}_\text{valid}|$, and $|\mathcal{N}_\text{test}| = |\mathcal{P}_\text{test}|$.

\emph{Relevance-based sampling} relies on random generation of negative examples 
involving predicates  from the heads of rules $\mathcal R$, and 
entities from the triples of $\mathcal K$ contributing to the support of these rules (i.e., matching the rules' bodies); additionally, it is ensured that the generated 
examples are truly negative---that is, not mentioned in any of the positive sets. 
Formally, let $\mathsf{Pred}_{\mathcal R}$ be the set of predicates in the heads of the rules in $\mathcal R$, $\mathsf{Const}_{\text{sup}}$ be all the 
constants from $\mathcal{K}$ occurring in the support of rules in $\mathcal R$, and
$\mathcal N_{\text{cand}}$ be the 
set of triples over $\mathsf{Pred}_{\mathcal R}$ and $\mathsf{Const}_{\text{sup}}$. Then, we take $\mathcal{N}_\text{train}$, $\mathcal{N}_\text{valid}$, and $\mathcal{N}_\text{test}$ as 
disjoint sets of the size specified above randomly 
sampled from $\mathcal N_{\text{cand}} \setminus \mathcal{P}_\text{all}$ without repetition. 
Note that this simple method is likely to ensure Requirement 2, 
because the predicates and constants of such negative examples are all involved in rule applications generating positive examples, and so it is likely that they represent rules that are similar to $\mathcal R$ but should not be learned.

\emph{Position-aware corruption} is a 
more fine-grained approach than relevance-based sampling. In particular, instead of sampling from all unseen triples
 in $\mathcal N_{\text{cand}}$, sampling is restricted to conclusion triples 
 corrupted with constants seen in a similar context. Formally, let $\mathcal{C}_\text{train}$, $\mathcal{C}_\text{valid}$, and $\mathcal{C}_\text{test}$ be the sets of conclusion triples in $\mathcal{P}_\text{train}$, $\mathcal{P}_\text{valid}$, and $\mathcal{P}_\text{test}$, respectively, for rules $\mathcal R$ in $\mathcal K$ (note that, 
 by construction, $\mathcal{C}_\text{valid} = \mathcal{P}_\text{valid}$ and $\mathcal{C}_\text{test} = \mathcal{P}_\text{test}$, but we give them different names for uniformity). Let
 $\mathcal{C}'_\text{train}$ be the set that contains 
 \begin{itemize}
 \item[--] each triple ${(e', \mathsf{type}, t) \notin \mathcal{P}_\text{all}}$ such that there exist triples ${(e, \mathsf{type}, t) \in \mathcal{C}_\text{train}}$ and ${(e', \mathsf{type}, t') \in \mathcal{P}_\text{all}}$; \item[--] each triple ${(s', R, o) \notin \mathcal{P}_\text{all}}$ such that there exist triples ${(s, R, o) \in \mathcal{C}_\text{train}}$ and ${(s', R, o') \in \mathcal{P}_\text{all}}$; and 
 \item[--]
each triple ${(s, R, o') \notin \mathcal{P}_\text{all}}$ such that there exist triples ${(s, R, o) \in \mathcal{C}_\text{train}}$ and ${(s', R, o') \in \mathcal{P}_\text{all}}$. 
\end{itemize}
Moreover, let $\mathcal{C}'_\text{valid}$ and $\mathcal{C}'_\text{test}$ be constructed in the same way from $\mathcal{C}_\text{valid}$ and $\mathcal{C}_\text{test}$, respectively.
Finally, we take $\mathcal{N}_\text{train}$, $\mathcal{N}_\text{valid}$, and $\mathcal{N}_\text{test}$ as sets of the sizes as specified above randomly sampled from $\mathcal{C}'_\text{train}$, $\mathcal{C'}_\text{valid} \setminus \mathcal{N}_\text{train}$, and $\mathcal{C}'_\text{test} \setminus (\mathcal{N}_\text{train} \cup \mathcal{N}_\text{valid})$,
 respectively, without repetition (note that the set difference is taken to avoid data leakage). 
 This simple method makes it 
 even more likely that the predicates and constants represent undesired rules  similar to $\mathcal R$, and hence
 makes further progress towards ensuring our Requirement 2 for inferential benchmarks.

\emph{Query-guided sampling} refines position-aware corruption and puts more emphasis on 
preventing systems from learning simple rules that entail rules in $\mathcal R$ (e.g., Rule \eqref{eq:ex_rule} follows from Rule~\eqref{eq:ex_rule1}  and so, assuming that~\eqref{eq:ex_rule} is in $\mathcal R$, we should ensure that~\eqref{eq:ex_rule1} is not learned, unless it logically follows from other rules in $\mathcal R$). Since there is 
usually an infinite number of rules logically entailing another rule, 
we take a pragmatic approach and, for the purpose of generating negative examples,
concentrate on rules 
obtained from rules in $\mathcal{R}$ by taking a subset of their body atoms. 
To cover rules that do not have any sub-rules entailing them, including those
with a single body atom, we also generate negative examples using the position-aware corruption method. 
Formally, let $\mathcal R^-$ be the set of rules 
obtained from rules in $\mathcal R$ by removing one or more body atoms, and let $\mathcal R_\text{complex}$ be the rules in $\mathcal R$ that contribute to this process. 
Then, let $\mathcal C^-_\text{train}$, $\mathcal C^-_\text{valid}$, and $\mathcal C^-_\text{test}$ be a split of the set $\mathcal T_{\mathcal R^-} (\mathcal K) \setminus \mathcal P_\text{all}$ 
with the same ratio as we used in Section~\ref{sec:rule-application}. Finally, we take $\mathcal{N}_\text{train}$, $\mathcal{N}_\text{valid}$,
 and $\mathcal{N}_\text{test}$ as sets of the sizes as specified 
 above constructed as follows: up to $|\mathcal R_\text{complex}| / |\mathcal R|$ fraction 
 of needed triples is sampled from $\mathcal C^-_\text{train}$, $\mathcal C^-_\text{valid}$, 
 and $\mathcal C^-_\text{test}$ without repetition, and the rest is 
 sampled from $\mathcal{C}'_\text{train}$, $\mathcal{C'}_\text{valid} \setminus \mathcal{N}_\text{train}$, 
 and $\mathcal{C}'_\text{test} \setminus (\mathcal{N}_\text{train} \cup \mathcal{N}_\text{valid})$, respectively, 
 where $\mathcal{C}'_\text{train}$, $\mathcal{C'}_\text{valid} $, and $\mathcal{C}'_\text{test}$ 
 are defined as in the position-aware case (note that `up to' in the first sampling 
 is essential because $\mathcal C^-_\text{train}$, $\mathcal C^-_\text{valid}$, and $\mathcal C^-_\text{test}$ may not have sufficiently many triples;
  however, the sampling here adheres as much as possible to the specified bound).

%% file: evaluation.tex
\section{Benchmarks}\label{sec:benchmarks}
   
Following our methodology in Section~\ref{sec:procedure}, we have constructed a 
suite of 37 benchmarks built upon three KGs: those underpinning 
the standard benchmarks FB15K-237~\cite{DBLP:conf/acl-cvsc/ToutanovaC15} and WN18RR~\cite{DBLP:conf/aaai/DettmersMS018}, denoted as
$\mathcal K_\text{fb}$ and $\mathcal K_\text{wn}$,  and the synthetic KG LUBM(1,0) ~\cite{DBLP:journals/ws/GuoPH05}, denoted as
$\mathcal K_\text{lubm}$.
We used RDFox~\cite{DBLP:conf/semweb/NenovPMHWB15} SPARQL engine to compute $\mathcal T_{\mathcal R}(\mathcal K)$ triples and in other similar cases.

Each benchmark based on $\mathcal K_\text{fb}$ and $\mathcal K_\text{wn}$ aims
to test systems' ability to learn a single inference pattern, and we concentrate on the 
patterns in Table~\ref{tab:patterns}. 
The intersection pattern on $\mathcal K_\text{wn}$ does not give a large-enough number of positive examples, so we omitted this case; we also
note that query-guided
negative sampling is relevant only to the
intersection, triangle, and diamond patterns.
So we constructed
31
benchmarks, and we use the notation
LogInfer-$X_{Z}^{Y}$, where $X \in \{\text{FB}, \text{WN}\}$ specifies the KG, $\mathcal K_\text{fb}$ or $\mathcal K_\text{wn}$, ${Y \in \{\text{sym}, \text{inver}, \text{hier}, \text{comp}, \text{inter}, \text{trian}, \text{diam}\}}$ specifies 
the pattern in Table~\ref{tab:patterns}, and $Z \in \{\text{rb}, \text{pa}, \text{qg}\}$ specifies the negative example generation method (with $Y = \text{inter}$ and $Z = \text{qg}$ applicable to only some cases as described above).
We
 used  8:1:1 as splitting ratio, and took numbers $k_1$ and $k_2$ as specified in Table~\ref{tab:benchmarks} to ensure appropriate size and variety.

The benchmarks based on $\mathcal K_\text{lubm}$ rely on 
the $107$ rules provided by Nenov et al.~\shortcite{DBLP:conf/semweb/NenovPMHWB15}. These rules are designed
to have large support, and hence are well-suited for generating sets of positive examples; furthermore,
they instantiate a wide range of inference patterns, including
symmetry, hierarchy, inversion, and composition.
We consider two variants of these benchmarks: one that uses all the 
rules, 
and one that uses only the $19$ rules not mentioning any types.
The latter is justified by the fact that many KG completion systems
do not have special treatment for type triples and consider them as regular triples over the $\mathsf{type}$ `relation';
this leads to significantly poorer performance in comparison to approaches with dedicated care of type triples~\cite{xie2016representation}. 
Overall, we constructed $6$ benchmarks based on LUBM denoted 
as LogInfer-$\text{LUBM}_{Z}^{Y}$, where $Y \in \{\text{all}, \text{no-type}\}$ specifies whether the rules
with types are included or not, and $Z \in \{\text{rb}, \text{pa}, \text{qg}\}$ specifies the negative example generation method. We 
used the ratio 8:1:1, and $k_1$ and $k_2$ as in Table~\ref{tab:benchmarks}.

The statistics of constructed benchmarks are summarised in 
Table~\ref{tab:benchmarks}. The 
benchmarks themselves and the
accompanying documentation 
are available online.\footnote{\href{https://github.com/shuwen-liu-ox/LogInfer}{https://github.com/shuwen-liu-ox/LogInfer}} 

\input{table_benchmarks}

\section{Evaluation} \label{sec:evaluation}

We have evaluated a representative sample of eight state-of-the-art
KG completion systems on the benchmarks described 
in Section \ref{sec:benchmarks}. 

\subsection{Systems and Training}
The evaluated systems can be divided into the following three categories:
\begin{enumerate}
    \item \emph{embedding-based} methods, which include TransE~\cite{DBLP:conf/nips/BordesUGWY13}, RotatE~\cite{DBLP:conf/iclr/SunDNT19},
    ComplEx~\cite{DBLP:conf/icml/TrouillonWRGB16}, 
    DistMult~\cite{DBLP:journals/corr/YangYHGD14a}, and
    BoxE~\cite{DBLP:conf/nips/AbboudCLS20};
    \item \emph{GNN-based} methods, including R-GCN~\cite{DBLP:conf/esws/SchlichtkrullKB18}; and
    \item \emph{rule mining} methods, including AnyBURL~\cite{DBLP:conf/ijcai/MeilickeCRS19} and
    RuleN~\cite{DBLP:conf/semweb/MeilickeFWRGS18}.
 \end{enumerate}
In the case of TransE, RotatE, ComplEx, and DistMult, we used the implementations provided by
Sun et al. \shortcite{DBLP:conf/iclr/SunDNT19}; for the other systems, we used
the implementations provided by the respective authors.
Additionally, we have implemented a simple baseline SimpBL, which
predicts a test triple $(a, b, c)$ as true if and only if 
$\mathcal{P}_\text{train}$ contains a triple involving both 
$a$ and $b$, and a triple involving both $b$ and $c$.

All the evaluated state-of-the-art systems 
provide confidence values in $[0, 1]$ for each prediction;
the threshold needed for computing classification-based metrics is therefore
considered a hyperparameter optimised during validation.
In contrast, SimpBL outputs Boolean 
values and hence metrics relying on confidence predictions 
are not applicable.

\subsection{Results}
\label{results}
\input{table_result_fb_wn}

\input{table_result_lubm_one_step}

We have evaluated all systems on all the benchmarks described
in Section~\ref{sec:benchmarks}. However, due to the large number of 
benchmarks and metrics, we
report only a representative selection of the obtained results.
In particular, we report (ROC) AUC, F1, precision, recall, C-MRR and R-MRR, 
and concentrate on query-guided method 
where applicable and position-aware corruption in the remaining cases.
The results for LogInfer-FB$_{Z}^{Y}$ and LogInfer-WN$_{Z}^{Y}$, where
 $Z = \text{pa}$ for $Y \in \{\text{sym, inver, hier, comp}\}$, and  $Z = \text{qg}$ for $Y \in \{\text{inter, trian, diam}\}$ are 
 given in
 Table~\ref{tab:result-fb-wn}; in turn, 
 the results for LogInfer-LUBM$_\text{qg}^\text{all}$ and LogInfer-LUBM$_\text{qg}^\text{no\_type}$ are in Table~\ref{tab:result-lubm-one-step}.
Our results can be summarised as follows.
\begin{enumerate}
  \item  All systems clearly and consistently outperformed our simple baseline on all benchmarks, with
  BoxE and RotatE  outperforming all other embedding-based methods.
  \item Systems' performance was  significantly better across the board 
for simple inference patterns (i.e., symmetry, inversion, and hierarchy) than for more
complex patterns involving conjunctions and inequalities in bodies (i.e., composition, triangle, and
diamond).
\item Rule-based systems exhibited 
better performance than the other systems on simple patterns; however, the gain is not significant on complex patterns. This can
be attributed to the fact that rule-based systems, on the one hand, mine rules based 
on the support of their bodies in the training set but, on the other hand, do not exploit the 
negative training examples, which penalise systems for learning unintended rules.
\item Some of our results are well-aligned with the theoretical findings on
the expressive power of KG completion models; for instance, 
TransE and DistMult performed  poorly on patterns that they cannot
capture theoretically (namely symmetry and inversion, respectively).
In contrast, other models achieved good performance 
on patterns that they cannot theoretically  capture;
for example, RotatE cannot capture hierarchy, but showed 
strong performance on LogicInfer-WN$^\text{hier}_\text{pa}$ and LogicInfer-FB$^\text{hier}_\text{pa}$. A possible reason is that RotatE captured a more specific pattern: it can capture $(x, R, y) 
        \rightarrow (x, S, y)$ provided the embeddings of $R$ and $S$ coincide, in which case $(x, S, y) 
        \rightarrow (x, R, y)$ also holds. By making predictions using $(x, R, y) 
        \leftrightarrow (x, S, y)$, RotateE may perform well. 
\item Relative performance varied across different metrics; for instance, R-GCN generally outperforms other models on R-MRR, but fares worse on C-MRR;
this can be explained by the design of R-GCN, which learns relation-specific parameters and so is adept at distinguishing relations.
\end{enumerate}

\subsection{Impact of Negative Example Generation}

We have studied the impact of the 
different negative example generation methods on
LogInfer-FB$^\text{sym}_Z$ for $Z \in \{\text{rc}, \text{rb}, \text{pa}\}$
and LogInfer-FB$^\text{inter}_Z$ for $Z \in \{\text{rc}, \text{rb}, \text{pa}, \text{qg}\}$; here, $Z = \text{rc}$ corresponds to 
the conventional random object corruption method where, for each triple $(s, R, o)$ in $\mathcal{P}_\text{train}$, $\mathcal{P}_\text{valid}$, and $\mathcal{P}_\text{test}$,
the corresponding $\mathcal{N}_\text{train}$, $\mathcal{N}_\text{valid}$, and $\mathcal{N}_\text{test}$ contains a triple $(s, R, o') \notin \mathcal{P}_\text{all}$ for $o'$ randomly sampled from $\mathsf{Consts}(\mathcal{K})$ in a way that the negative sets remain disjoint.

Table~\ref{tab:result-neg} summarises our
 results for the
 AUC metric.
As we can see, systems achieved high scores whenever the conventional random corruption method was used
($90.9\%$ on average for symmetry and 
$96\%$ for intersection); this aligns with
previous criticism indicating that such negative examples are
trivial to recognise. 
In contrast, performance consistently degraded as we adopted
increasingly fine-grained methods:  average scores for relevance-aware corruption
drop to $88\%$ (symmetry) and $91.7\%$ (intersection), 
and further degrade to $85.5\%$ and  $89.3\%$ respectively
for position-aware corruption.
In turn, average performance 
drops to $85.6\%$
for the intersection
pattern for the query-guided approach. 

These results support the effectiveness of our proposed methods in 
generating challenging negative examples. 
Overall, our findings indicate 
that classification performance heavily relies on the choice of negative examples, and thus 
highlights the importance of  devising
carefully designed methods for this choice in KG completion benchmarks.

\input{table_result_neg}

\subsection{Analysis of Extracted Rules}\label{sec:rule-analysis}
AnyBURL and RuleN explicitly construct 
sets of Datalog rules; therefore, we can compare the sets of rules
returned by these systems with those in the benchmarks, $\mathcal R$, without relying
on specific test examples. 
To this end, we considered LogInfer-FB$^Y_Z$ for all patterns $Y$ as representative
benchmarks (note that the value of $Z$ is irrelevant since
rule-based systems do not exploit negative training examples), 
and computed the following metrics:
\begin{enumerate}
  \item the percentage $\epsilon_\text{ent}$ of benchmark rules $\mathcal R$ logically entailed by the rule sets generated by the systems; and
  \item the percentage $\epsilon_\text{cont}$ of rules in $\mathcal R$ syntactically generated (and therefore also entailed) by the systems
  up to variable renaming and permutations of body atoms. 
\end{enumerate}
 It is worth emphasising that 
 these systems generate very large rule sets
 because rules are extracted by identifying dependencies in
 the training data, which includes the original KG; as a result,
 many generated rules are irrelevant to $\mathcal R$. 
 Therefore, computing the percentages of extracted rules that are entailed 
 by the benchmark rules is not very meaningful.

The results  are summarised in
Table~\ref{tab:result-rule}. Over 95\% benchmark rules 
corresponding to simple patterns 
are  syntactically included 
in the output rule sets of AnyBURL and RuleN; this aligns
with their superior performance on these patterns. 

We further analysed rules for complex patterns.
AnyBURL and RuleN only generate Datalog rules 
of certain syntactic form; for instance, AnyBURL and RuleN
cannot syntactically output rules for
intersection, triangle, or diamond patterns. A very large proportion of these
rules are, however, entailed by the systems' output rule sets; this can be explained
by the fact that the systems are generating simpler rules instead. 
To verify this, we have focused on the triangle pattern and have generated
all the rules $\mathcal{R}^-$ obtained by selecting subsets of body atoms in the
benchmark rules for the pattern 
as described in Section \ref{sec:negative-sampling}.
We could verify that 
 $99.5\%$ of these rules
 were entailed by the rules extracted by AnyBURL, and $89.5 \%$ of
 them were entailed by the rules extracted by RuleN.
This explains the performance drop observed for rule-based systems
on complex patterns and further supports the 
identified need for better negative example generation methods that
penalise systems accordingly.
\input{table_result_rule.tex}

%% file: table_benchmarks.tex
\begin{table}[t]
\begin{small}
\centering
\renewcommand\arraystretch{1}
\setlength{\tabcolsep}{0.9mm}{
\begin{tabular}{ccrrrrrr}
\toprule
   & $Y$ & $k_1$ & $k_2$ & $\vert\mathcal{K}\vert$~~~~ & $\vert\mathcal{P}_\text{train}\vert$~~ & $\vert\mathcal{P}_\text{valid}\vert$ & $\vert\mathcal{P}_\text{test}\vert$\\
\midrule
\multirow{7}{*}{\begin{tabular}{r}\!LogInfer-~~ \\ FB$^Y_Z$\end{tabular}} 
& \small{sym} 
  & 50 & 200 & 310,116 & 318,116 & 1,000 & 1,000\\
 & \small{inver} 
 & 200 & 200 & 310,116 & 341,603 & 3,928  & 3,946 \\
 & \small{hier} 
 & 200 & 200 & 310,116 & 341,477 & 3,915  & 3,934 \\
 & \small{inter} 
 & 200 & 200 & 310,116 & 333,478 & 2,908  & 3,126 \\
 & \small{comp} 
 & 200 & 200 & 310,116 & 341,829 & 3,964 & 3,966 \\
 & \small{trian} 
 & 200 & 200 & 310,116 & 341,974 & 3,975  & 3,975 \\
 & \small{diam} 
 & 200 & 200 & 310,116 & 340,625 & 3,803  & 3,822 \\
\midrule
\multirow{6}{*}{\begin{tabular}{r}\!LogInfer-~~ \\ WN$^Y_Z$\end{tabular}} 
 & \small{sym} 
 & ~~~~5 & 2000 & ~~93,003 & 101,003 & 1,000 & 1,000\\
 & \small{inver} 
 & ~~~~5 & 2000 & ~~93,003 & 101,003 & 1,000 & 1,000\\
 & \small{hier} 
 & ~~~~5 & 2000 & ~~93,003 & 101,003 & 1,000 & 1,000\\
 & \small{comp} 
 & ~~20 & 2000 & ~~93,003 & 114,336 & 1,610  & 1,564 \\
 & \small{trian} 
 & ~~20 & 2000 & ~~93,003 & 100,800 & ~~~968  & ~~~977 \\
 & \small{diam} 
 & ~~20 & 2000 & ~~93,003 & 104,681 & 1,398  & 1,391 \\
\midrule
\small{\!LogInfer-~~~} & 
\small{all} 
 & 107 & 500 & 103,119 & 140,536 & 5,001 & 5,003\\
\small{~~~LUBM$^Y_Z$}
& \!\!\small{no-type}\!\!
 & ~~19 & 2500 & 103,119 & 140,540  & 4,999 & 5,002 \\
\bottomrule
\end{tabular}
}
\caption{Benchmark statistics, where each number applies for all relevant $Z$ (the sizes of the negative example sets $\vert\mathcal{N}_\text{train}\vert$, $\vert\mathcal{N}_\text{valid}\vert$, and $\vert\mathcal{N}_\text{test}\vert$ are the same as for positive examples)
}
\label{tab:benchmarks}
\end{small}
\end{table}

%% file: table_result_fb_wn.tex
\begin{table*}[!t]
\centering
\begin{small}
\setlength{\tabcolsep}{1.1mm}
\renewcommand\arraystretch{0.902}
\begin{tabular}{ccc|cccccc|cccccc}
\toprule
  &    &     &\multicolumn{6}{c|}{\small{LogInfer-FB}$_{Z}^{Y}$} & \multicolumn{6}{c}{\small{LogInfer-WN$_{Z}^{Y}$}}\\
$Y$ & $Z$& Model
& AUC & F1   & Precision    & Recall
& R-MRR & C-MRR
& AUC & F1    & Precision    & Recall
& R-MRR & C-MRR\\
\midrule
\multirow{9}{*}{sym} & \multirow{9}{*}{pa} & SimpBL   & - 
& 66.7 & 50.0      & \textbf{100.0}  & -     & -     & - 
& ~~30.5  & ~~31.8      & ~~29.4   & -     & -     \\
 & & TransE   & 47.0    & 56.2 & 45.9      & ~~72.3   & ~~9.1   & ~~3.4   & ~~98.8    & ~~98.8  & ~~98.6      & ~~99.0   & 45.1  & ~~31.3  \\
 & & RotatE   & 86.8    & 93.1 & 87.7      & ~~99.1   & 56.8  & 19.9  & ~~99.1    & ~~99.1  & ~~98.5      & ~~99.8   & 50.0  & ~~61.6  \\
 & & ComplEx  & 73.0    & 83.5 & 74.0      & ~~95.8   & 48.3  & 16.3  & ~~80.3    & ~~81.1  & ~~77.8      & ~~84.8   & 49.4  & ~~~~6.9   \\
 & & DistMult & 97.5    & 98.6 & 97.7      & ~~99.5   & \textbf{89.7}  & 41.7  & ~~82.3    & ~~83.1  & ~~79.5      & ~~87.0   & 50.4  & ~~15.1  \\
 & & BoxE     & 92.8    & 94.9 & 95.3      & ~~94.6   & 87.3  & 30.6  & \textbf{100.0}   & \textbf{100.0} & \textbf{100.0}     & \textbf{100.0}  & 53.6  & ~~\textbf{68.8}  \\
 & & R-GCN    & 92.8    & 96.2 & 92.8      & \textbf{100.0}  & 81.0  & 17.4  & ~~96.1    & ~~96.2  & ~~92.8      & \textbf{100.0}  & \textbf{81.0}  & ~~54.7  \\
 & & AnyBURL  & 96.0    & 97.9 & 96.0      & \textbf{100.0}  & 86.3  & 43.6  & ~~99.9    & ~~99.9  & ~~99.9      & \textbf{100.0}  & 53.5  & ~~63.7  \\
 & & RuleN    & \textbf{98.0}    & \textbf{99.0} & \textbf{98.0}      & \textbf{100.0}  & 86.6  & \textbf{43.7}  & ~~88.3    & ~~86.7  & \textbf{100.0}     & ~~76.5   & 53.8  & ~~51.1  \\
\midrule
\multirow{9}{*}{inver} & \multirow{9}{*}{pa} & SimpBL   & -
 & 36.2 & 38.2      & ~~34.4   & -     & -     & -   
   & ~~31.1  & ~~31.6      & ~~30.6   & -     & -     \\
 & & TransE   & 90.1    & 90.2 & 88.9      & ~~91.6   & 69.9  & 34.3 & ~~87.6  & ~~88.5 &~~82.3         & ~~95.6   & 48.5  & ~~18.7  \\
 & & RotatE   & 91.4    & 92.0 & 87.9      & ~~96.4   & 78.6  & 56.3  & ~~88.6   & ~~89.2  & ~~84.3      & ~~94.5   & 87.2  & ~~46.1  \\
 & & ComplEx  & 81.4    & 81.4 & 81.4      & ~~81.3   & 52.7  & 43.7  & ~~80.8    & ~~81.1  & ~~80.4      & ~~81.9   & 65.2  & ~~29.2   \\
 & & DistMult & 83.2    & 83.2 & 83.4      & ~~82.9   & 45.0  & 44.9  & ~~88.4    & ~~88.2  & ~~89.8      & ~~88.6   & 75.3  & ~~36.3  \\
 & & BoxE     & \textbf{94.2}    & 94.2 & 94.1      & ~~94.2   & 78.7  & 32.8  & ~~93.0   & ~~93.5  & ~~{92.8}      & ~~94.2   & \textbf{91.3}  & ~~53.2  \\
 & & R-GCN    & 84.3    & 84.3 & 83.7      & ~~84.8   & 56.2  & 35.7   & ~~96.9    & ~~96.9  & ~~\textbf{99.7}      & ~~94.2   & 87.9  & ~~65.6   \\
 & & AnyBURL  & 93.9    & \textbf{94.3} & 90.0      & ~~\textbf{98.9}   & \textbf{80.8}  & \textbf{56.9}  & ~~\textbf{99.8}    & ~~\textbf{99.9}  & ~~\textbf{99.7}      & \textbf{100.0}   & 90.3  & ~~\textbf{67.2}  \\
 & & RuleN    & 86.0    & 84.2 & \textbf{96.3}      & ~~74.8   & 62.9  & 41.9  & ~~85.8    & ~~83.5  & ~~99.6      & ~~71.9   & 73.1  & ~~42.0 \\
\midrule
\multirow{9}{*}{hier} & \multirow{9}{*}{pa} &  SimpBL   & - 
   & 36.1 & 39.5      & ~~33.2   & -     & -     & -
     & ~~31.8  & ~~35.7      & ~~28.6   & -     & -     \\
 & & TransE   & 90.1    & 90.4 & 87.9      & ~~93.0   & 29.6  & 17.2  & ~~99.4    & ~~99.4  & ~~99.3      & ~~99.5   & 33.4  & ~~27.8  \\
 & & RotatE   & 91.9    & 92.1 & 89.3      & ~~95.1   & 42.7  & 26.5  & ~~98.8    & ~~98.8  & ~~98.4      & ~~99.2   & 66.4  & ~~69.1  \\
 & & ComplEx  & 81.3    & 80.5 & 84.0      & ~~77.3   & 26.4  & 16.7  & ~~81.0    & ~~80.1  & ~~84.0      & ~~76.5   & 53.3  & ~~16.5  \\
 & & DistMult & 80.5    & 80.9 & 79.5      & ~~82.3   & 27.0  & 19.7  & ~~83.6    & ~~84.1  & ~~81.8      & ~~86.4   & 59.6  & ~~28.5  \\
 & & BoxE     &\textbf{94.9}    & \textbf{94.9} & 94.2      & ~~95.7   & 77.2  & 32.9  & ~~\textbf{99.9}    & ~~\textbf{99.9}  & ~~\textbf{99.9}      & ~~99.9   & 86.7  & ~~57.6  \\
 & & R-GCN    & 84.5    & 85.6 & 79.9      & ~~92.2   & 72.5  & 34.5  & ~~65.5    & ~~67.9  & ~~63.4      & ~~73.2   & \textbf{95.5}  & ~~61.1  \\
 & & AnyBURL  & 91.8    & 92.4 & 86.0      & ~~\textbf{99.9}   & \textbf{82.1}  & \textbf{56.2}  & ~~\textbf{99.9}    & ~~\textbf{99.9}  & ~~\textbf{99.9}      & \textbf{100.0}  & 94.4  & ~~\textbf{73.5}  \\
  && RuleN    & 85.3    & 83.4 & \textbf{96.4}      & ~~73.4   & 60.6  & 39.1  & ~~82.5    & ~~79.0  & ~~99.0      & ~~65.7   & 67.7  & ~~35.8  \\
 \midrule
\multirow{9}{*}{comp} & \multirow{9}{*}{pa} & SimpBL   & -
  & 25.8 & 30.9      & ~~22.2   & -     & -     & - 
   & ~~31.6  & ~~29.8      & ~~33.6   & -     & -     \\
  && TransE   & 86.9    & 87.1 & 79.9      & ~~95.7   & 13.9  & ~~8.2   & ~~\textbf{99.7}    & ~~\textbf{99.7}  & ~~\textbf{99.7}      & ~~99.7   & 32.4  & ~~49.9  \\
  && RotatE   & 88.1    & 89.0 & 83.5      & ~~95.2   & 24.6  & 16.3  & ~~99.0    & ~~99.0  & ~~99.4      & ~~98.6   & 45.2  & ~~84.2  \\
  && ComplEx  & 68.6    & 70.9 & 66.1      & ~~76.5   & ~~9.7   & ~~7.3   & ~~84.5    & ~~85.1  & ~~81.7      & ~~88.9   & 45.2  & ~~~~9.5   \\
  && DistMult & 70.4    & 72.6 & 67.6      & ~~78.4   & ~~9.8   & ~~8.2   & ~~95.4    & ~~95.3  & ~~98.2      & ~~92.5   & 50.9  & ~~64.7  \\
 & & BoxE     & 90.1    & 90.6 & 87.0      & ~~94.5   & 39.3  & 23.7  & ~~\textbf{99.7}    & ~~\textbf{99.7}  & ~~\textbf{99.7}      & ~~99.7   & 81.3  & ~~61.9  \\
  && R-GCN    & 73.6    & 74.9 & 71.3      & ~~79.0   & ~~8.2   & ~~9.8   & ~~75.2    & ~~76.6  & ~~72.6      & ~~81.1   & 41.9  & ~~~~9.0   \\
 & & AnyBURL  & \textbf{91.7}    & \textbf{92.3} & 86.3      & ~~\textbf{99.2}   & \textbf{39.4}  & \textbf{34.1}  & ~~99.3    & ~~99.3  & ~~98.7      & ~~\textbf{99.9}   & \textbf{92.9}  & ~~\textbf{93.4}  \\
 & & RuleN    & 86.5    & 85.1 & \textbf{94.3}      & ~~77.6   & 31.7  & 26.1  & ~~91.7    & ~~91.0  & ~~98.7      & ~~84.5   & 75.7  & ~~69.0  \\
\midrule
\multirow{9}{*}{inter}  & \multirow{9}{*}{qg}& SimpBL & - 
 & 65.5 & \textbf{90.8}      & ~~51.2   & -     & -     & -       & -     & -         & -      & -     & -     \\
  && TransE   & 87.1    & 87.3 & 86.2      & ~~88.4   & 20.9   & 44.6   & -       & -     & -         & -      & -     & -     \\
  && RotatE   & \textbf{89.3}    & \textbf{89.4} & 89.1      & ~~89.6   & 33.4   & \textbf{78.4}  & -       & -     & -         & -      & -     & -     \\
  && ComplEx  & 83.4    & 82.0 & 89.3      & ~~75.8   & 17.7   & 65.1   & -       & -     & -         & -      & -     & -     \\
  && DistMult & 85.7    & 86.1 & 83.6      & ~~88.7   & 20.3   & 63.2  & -       & -     & -         & -      & -     & -     \\
  && BoxE     & 85.3    & 88.9 & 85.3      & ~~\textbf{95.1}   & 60.7  & 40.5  & -       & -     & -         & -      & -     & -     \\
  && R-GCN    & 85.9    & 85.3 & 89.0      & ~~81.9   & 61.1   & 44.2  & -       & -     & -         & -      & -     & -     \\
  && AnyBURL  & 88.2    & 88.4 & 87.5      & ~~89.3   & \textbf{76.4}  & 71.0  & -       & -     & -         & -      & -     & -     \\
  && RuleN    & 80.2    & 78.1 & 87.5      & ~~70.5   & 58.9  & 52.8  & -       & -     & -         & -      & -     & -     \\
\midrule
\multirow{9}{*}{trian} & \multirow{9}{*}{qg} & SimpBL   & -
   & 52.2 & 49.4      & ~~55.2   & -     & -     & -
     & ~~56.3  & ~~87.7      & ~~41.5   & -     & -     \\
  && TransE   & 91.7    & 92.1 & 88.2      & ~~96.3   & 23.5   & 18.9   & ~~80.2    & ~~79.2  & ~~88.9      & ~~71.4   & 36.7  & ~~48.8   \\
  && RotatE   & \textbf{93.1}    & \textbf{93.1}& \textbf{90.8}      & ~~95.6   & 31.0  & \textbf{31.5}  & ~~89.1    & ~~88.5  & ~~\textbf{93.4}      & ~~84.1   & 43.0  & ~~48.2  \\
  && ComplEx  & 77.0    & 77.9 & 74.9      & ~~81.2   & 23.5   & 18.9   & ~~74.4    & ~~70.8  & ~~82.2      & ~~62.2   & 27.1  & ~~12.6  \\
  && DistMult & 81.2    & 81.9 & 78.7      & ~~85.4   & 20.3   & 25.4   & ~~83.8    & ~~82.3  & ~~90.3      & ~~75.6   & 30.7  & ~~31.8  \\
 & & BoxE     & 82.7    & 85.0 & 75.0      & ~~\textbf{98.0}   & \textbf{36.3}  & 21.1  & ~~58.9    & ~~70.0  & ~~54.9      & ~~\textbf{96.3}   & 69.5  & ~~55.5  \\
  && R-GCN    & 85.3    & 87.0 & 78.6      & ~~97.3   & 35.1  & 22.5  & ~~86.6    & ~~85.7  & ~~91.9      & ~~80.2   & 75.8  & ~~71.3   \\
 & & AnyBURL  & 86.4    & 87.1 & 83.3      & ~~91.2   & 33.5  & 26.6  & ~~\textbf{89.7}    & ~~\textbf{89.8}  & ~~89.2      & ~~90.4   & \textbf{86.3}  & ~~\textbf{81.7}  \\
  && RuleN    & 84.8    & 84.5 & 86.6      & ~~82.5   & 27.4  & 19.8  & ~~82.5    & ~~80.8  & ~~89.3      & ~~73.9   & 79.5  & ~~64.1  \\
 \midrule
\multirow{9}{*}{diam}& \multirow{9}{*}{qg} & SimpBL   & - 
  & 60.2 & 67.6      & ~~54.2   & -     & -     & -   
    & ~~48.7  & ~~69.3      & ~~37.5   & -     & -     \\
  && TransE   & 77.9    & 80.4 & 72.1      & ~~91.0   & 20.3  & 13.8  & ~~52.2    & ~~67.5  & ~~51.1      & ~~99.5   & 23.2  & ~~26.8  \\
  && RotatE   & 83.4    & 84.9 & 77.9      & ~~93.2   & 25.5  & 21.4  & ~~77.5    & ~~80.2  & ~~71.8      & ~~90.8   & 53.4  & ~~71.6  \\
  && ComplEx  & 64.2    & 70.3 & 60.1      & ~~84.8   & ~~8.4  & ~~6.2  & ~~55.0    & ~~67.0  & ~~52.8      & ~~91.6   & 35.7  & ~~~~1.6   \\
  && DistMult & 77.6    & 79.6 & 73.0      & ~~87.4   & 11.9  & 15.5  & ~~69.3    & ~~74.4  & ~~63.8      & ~~89.1   & 45.6  & ~~23.4  \\
  && BoxE     & 77.5   & 80.5 & 71.2      & ~~92.5   & 17.3 & 16.8  & ~~72.8    & ~~76.5  & ~~67.5      & ~~88.2   & 80.5  & ~~58.8  \\
  && R-GCN    & \textbf{86.6}    & \textbf{87.3} & \textbf{83.1}      & ~~92.0   & 20.5   & 18.2  & ~~\textbf{94.5}    & ~~\textbf{94.4 } & ~~\textbf{95.3}      & ~~\textbf{93.5}   & \textbf{89.6}   & ~~\textbf{70.9}   \\
  && AnyBURL  & 67.8    & 73.3 & 57.9      & ~~\textbf{99.9}   & \textbf{39.5}  & \textbf{35.2}  & ~~73.8    & ~~77.2  & ~~63.9      & ~~97.5   & 87.3  & ~~61.5  \\
  && RuleN    & 69.6    & 71.8 & 62.1      & ~~84.9   & 30.2  & 24.1  & ~~61.4    & ~~66.9  & ~~55.5      & ~~84.5   & 61.8  & ~~44.7 \\
\bottomrule
\end{tabular}
\caption{Evaluation results on LogInfer-FB$_{Z}^{Y}$ and LogInfer-WN$_{Z}^{Y}$ (in \%)}
\label{tab:result-fb-wn}
\end{small}
\end{table*}

%% file: table_result_lubm_one_step.tex
\begin{table}[!t]
\begin{small}
\centering
\setlength{\tabcolsep}{0.7mm}
\renewcommand\arraystretch{0.93}
\begin{tabular}{cc|rrrrrr}
\toprule
  &        &\multicolumn{6}{c}{LogInfer-LUBM$_\text{qg}^Y$\;}\\
$Y$ & Model
& AUC & F1~   & Precision    & Recall
& R-MRR & C-MRR\\
\midrule
\multirow{9}{*}{all} & SimpBL   & -~~
  & 64.3 & 49.2~~      & 92.8~   & -~~~     & -~~~     \\
 & TransE   & 96.3    & 96.3 & 95.6~~      & 97.0~   & 75.5~  & 26.8~  \\
 & RotatE   & 96.3    & 96.5 & 95.8~~      & 97.3~   & 96.7~  & 49.4~  \\
 & ComplEx  & 89.7    & 89.4 & 93.4~~      & 85.8~   & 84.2~  & 58.9~  \\
 & DistMult & 92.7    & 92.4 & 95.9~~      & 89.1~   & 86.9~  & \textbf{62.5}~  \\
 & BoxE     & 97.4    & 97.4 & 95.7~~      & 99.3~   & \textbf{100.0}~  & 12.5~  \\
 & R-GCN    & 94.1    & 94.1 & 94.1~~      & 94.0~   & 82.7~  & 34.9~  \\
 & AnyBURL  & \textbf{98.1}    & \textbf{98.4} & \textbf{97.0}~~      & \textbf{99.9}~   & \textbf{100.0}~  & 22.0~  \\
 & RuleN    & 88.3    & 91.2 & 95.0~~      & 87.6~   & 89.0~  & 18.1~  \\
 \midrule
& SimpBL  & -~~ 
  & 62.4 & 48.2~~      & 88.8~   & -~~~     & -~~~     \\
 & TransE   & 98.2  & 98.3 & 96.7~~      & \textbf{99.9}~   & 97.2~  & 45.7~  \\
 & RotatE   & 98.5  & 98.4  & 97.1~~      & 99.8~   & 97.6~  & 55.1~  \\
 no$\_$& ComplEx  & 89.9    & 89.3 & 94.6~~      & 84.6~   & 86.3~  & 49.7~  \\
 type & DistMult & 93.3    & 93.1 & 95.4~~      & 90.9~   & 84.5~  & 63.2~  \\
 & BoxE     & 99.3    & 99.3 & 98.8~~      & 99.8~   & \textbf{100.0}~  & 10.1~  \\
 & R-GCN    & 94.7    & 95.2 & 93.9~~      & 96.6~   & 81.5~  & 33.2~  \\
 & AnyBURL  & \textbf{99.9}    & \textbf{99.9} & \textbf{100.0}~~      & 99.8~   & \textbf{100.0}~  & 83.5~  \\
 & RuleN    & 92.4    & 96.5 & 94.9~~      & 98.2~   & 98.5~  & \textbf{98.2}~ \\
\bottomrule
\end{tabular}
\caption{Evaluation results on LogInfer-LUBM$_\text{qg}^Y$ (in \%)}
\label{tab:result-lubm-one-step}
\end{small}
\end{table}

%% file: table_result_neg.tex
\begin{table}[!t]
\begin{center}
\begin{small}
\setlength{\tabcolsep}{1.8mm}
\renewcommand\arraystretch{0.93}
\begin{tabular}{cccc|cccc}
\toprule
        &\multicolumn{3}{c|}{\small{LogInfer-FB$^\text{sym}_Z$}} & \multicolumn{3}{c}{\small{LogInfer-FB$^\text{inter}_Z$}}\\
~~~~~~~~$Z$ & rc & rb & pa & rc & rb & pa & qg \\
\midrule
TransE               & 62.6       & 56.7      & 47.0     & \textbf{98.9}       & 93.9     & 94.4  & 87.1  \\
RotatE               & 95.7       & 91.0      & 86.8     & 98.7       & \textbf{95.5}   & 93.8   & \textbf{89.3}  \\
ComplEx              & 79.7       & 78.5      & 73.0     & 88.8       & 86.9    & 81.1 &   83.4  \\
DistMult             & \textbf{99.8}       & \textbf{99.4}      & 97.5     & 97.0       & 92.9   &84.9&    85.7  \\
BoxE                 & 94.0       & 89.1      & 92.8     & 97.5       & 95.1  &   \textbf{95.9}&   85.3  \\
R-GCN                & 98.4       & 96.8      & 92.8     & 96.6       & 93.1    &87.4&   85.9  \\
AnyBURL              & 98.1       & 96.7      & 96.0    & 97.5      &91.4   &90.8   & 88.2  \\
RuleN                & 98.6       & 95.8      & \textbf{98.0}     & 92.8       & 84.7  &85.8    & 80.2 \\
\bottomrule
\end{tabular}
\caption{AUC scores with various negative example generation methods for classification-based evaluation (in $\%$)}
\label{tab:result-neg}
\end{small}
\end{center}
\end{table}

%% file: table_result_rule.tex
\begin{table}[!t]
\centering
\begin{small}
\renewcommand\arraystretch{0.93}
\setlength{\tabcolsep}{1.1mm}
\begin{tabular}{ccccccccc}
\toprule
& ~~~~~~~~~~~~~$Y$ & sym & inver & hier & comp & inter & trian & diam \\
\midrule
\multirow{2}{*}{$\epsilon_\text{ent}$}  & AnyBURL    & \textbf{100.0}      & \textbf{98.5}  & \textbf{99.0}   & \textbf{95.0}      & 90.5      & \textbf{99.5} & \textbf{89.5}\\
 & RuleN  &    ~~98.0      & 93.5 & \textbf{99.0}  & 92.0  & \textbf{97.5} & 94.0 & 73.5  \\
 \midrule
 \multirow{2}{*}{$\epsilon_\text{cont}$}  & AnyBURL     & 
 ~~96.0   &  \textbf{96.5} & \textbf{95.5}    & \textbf{34.0}     & ~~0.0       & ~~0.0  & ~~0.0 \\
 & RuleN  &  ~~\textbf{98.0}   & 70.5   & 69.5  & 23.0  & ~~0.0 & ~~0.0 & ~~0.0  \\
\bottomrule
\end{tabular}
\caption{Results for logical entailment on LogInfer-FB$^Y_Z$ (in \%)}
\label{tab:result-rule}
\end{small}
\end{table}

%% file: conclusion.tex
\section{Conclusion}
In this paper, we have presented a novel approach for generating challenging inferential KG completion benchmarks. 
On the one hand, our approach ensures that models are exposed during training to both premise and conclusion triples
for selected rules, and that triples in the test set have supporting evidence in the training set.
On the other hand, our methods for generating negative examples ensure that models
are penalised for learning unintended rules and yield examples that are both relevant and challenging to
classify. Our findings highlight the gaps  between theoretical and empirical results 
concerning models’ ability to capture inference patterns and open the door to 
future investigation.   